\documentclass[12pt, final]{l4dc2023}

\usepackage[frozencache,cachedir=.]{minted}
\usepackage{acronym}
\usepackage{bm}
\usepackage{hyperref}

\acrodef{DMD}{Dynamic Mode Decomposition}
\acrodef{ANAE}{Average Normalized Absolute Error}
\acrodefplural{CPS}[CPS]{cyber-physical systems}
\acrodef{UAV}{Unmanned Aerial Vehicle}
\acrodef{SwRI}{Southwest Research Institute}
\acrodef{SVD}{Singular Value Decomposition}
\acrodef{MLP}{multi-layer perceptron}
\acrodef{LRAN}{Linearly Recurrent Autoencoder Network}
\acrodef{MSE}{Mean Squared Error}
\acrodef{CFD}{computational fluid dynamics}
\acrodef{NN}{neural network}

\newcommand{\code}[1]{\texttt{#1}}
\newcommand{\package}{DLKoopman}
\newcommand{\SP}{\code{StatePred}}
\newcommand{\TP}{\code{TrajPred}}

\title[DLKoopman: A deep learning software package for Koopman theory]{DLKoopman: A deep learning software package for Koopman theory}
\usepackage{times}

\author{%
\Name{Sourya Dey} \Email{sourya@galois.com}\\
\Name{Eric William Davis} \Email{ewd@galois.com}\\
\addr 421 SW 6th Avenue \#300, Portland, OR 97204, USA%
}

\begin{document}
\sloppy

\maketitle

\begin{abstract}%
We present DLKoopman -- a software package for Koopman theory that uses deep learning to learn an encoding of a nonlinear dynamical system into a linear space, while simultaneously learning the linear dynamics. While several previous efforts have either restricted the ability to learn encodings, or been bespoke efforts designed for specific systems, DLKoopman is a generalized tool that can be applied to data-driven learning and analysis of any dynamical system. It can either be trained on data from individual states (snapshots) of a system and used to predict its unknown states, or trained on data from trajectories of a system and used to predict unknown trajectories for new initial states. DLKoopman is available on the Python Package Index (PyPI) as `dlkoopman', and includes extensive documentation and tutorials. Additional contributions of the package include a novel metric called Average Normalized Absolute Error for evaluating performance, and a ready-to-use hyperparameter search module for improving performance.
\end{abstract}


\section{Introduction}\label{sec:intro}
The abundance of data, combined with the rise in computational power, has enabled the creation of increasingly powerful machine learning systems to model and predict the world around us. In particular, \emph{deep learning} -- i.e. machine learning implemented via \acp{NN} comprising multiple layers -- has become extremely popular due to its ability to `intelligently' learn the rules of any system from its available data. One such data-driven application of deep learning is to learn the dynamics of a system purely from its states, or snapshots. This is especially useful in situations where the exact rules governing the system are prohibitively hard and/or time-consuming to understand and analyze, whereas data can be easily and plentifully collected from the system.

\Ac{DMD} is a technique to analyze nonlinear systems by approximating them using linear dynamics, i.e. \emph{linearizing} them. However, an arbitrary nonlinear system will, in general, be poorly suited to linearization. \emph{Koopman theory}\footnote{Also known as Koopman operator theory, or just Koopman operator.}, first introduced in \cite{Koopman1931}, overcomes this limitation by extending \ac{DMD} to encode states of a nonlinear system into \emph{observables} in a different domain, performing linearization in this \emph{encoded domain}, then decoding back into the original input domain. This domain shift is critical to improving the fidelity of the linear model, leading to significantly better approximations. The obtained linear model is incredibly powerful since linear techniques can be used to easily \emph{predict unknown states} of the system, as well as \emph{predict entire trajectories} of how the system behaves starting from new initial conditions.

A key challenge with implementing Koopman theory is finding an encoding for the original input domain into a linearizable domain. Particularly suited to this task is the \emph{autoencoder} deep learning architecture, which consists of an encoder \ac{NN} learning to convert input data to an encoded domain suited to linear approximations, connected to a decoder \ac{NN} simultaneously learning the inverse function of the encoder so as to convert back to the original input domain.

Our core contribution in this paper is introducing \package{} -- an open-source Python package to implement Koopman theory. It can be installed via \code{pip install dlkoopman}. To the best of our knowledge, we are the first effort to create a software package for Koopman theory that is a) \emph{general and reusable}, in the sense that it can operate on data from any system and perform state prediction, as well as trajectory prediction, and b) \emph{complete}, in the sense that it uses a deep learning pipeline that includes learning both -- the encoding into a linearizable domain, and the corresponding linear dynamics. \package{} bridges the gap between two schools of prior work -- a) software packages that use \ac{DMD} without learning an encoding into a domain that ensures good linearization, and b) efforts that use deep learning for the encoding, but lack general software tooling and are usually restricted to trajectory prediction. 

We make two additional contributions via the package. Firstly, we introduce a novel error function -- \emph{\ac{ANAE}} -- which is useful for quantifying performance and comparing different models. Secondly, we include a ready-to-use \emph{hyperparameter search module} that can provide significant performance gains when performing Koopman approximations.

\subsection{Mathematical background}\label{sec:intro-math}
The mathematical details relevant to this paper are in \cite{Dey2022_dmdkoopReport}, which we summarize here. For a more extensive mathematical treatment, the reader can refer to several sources such as \cite{DMDbook,Tu2014,Williams2015}. Suppose a system is described as $\frac{d\bm{x}}{di} = f\left(\bm{x}(i)\right)$, where $\bm{x}$ is the (generally multi-dimensional) state of the system indexed by $i$, which may, but need not necessarily, be time. The system can be sampled to obtain its states at various discrete indexes, which can be described as $\bm{x}_{i+1} = F\left(\bm{x}_i\right)$. Here, $f(\cdot)$ and $F(\cdot)$ are (generally nonlinear) functions encapsulating the dynamics of how the system evolves. The first step in applying Koopman theory is to transform the original $\bm{x}$ domain into an encoded domain $\bm{y}$ using an encoder $g(\cdot)$:
\begin{align}
\bm{y} = g\left(\bm{x}\right) \label{eq:koopman-encoder}
\end{align}

State evolution in the encoded domain is linear, i.e. $\frac{d\bm{y}}{di} = \bm{\mathcal{K}}\bm{y}(i)$, where $\bm{\mathcal{K}}$ is the Koopman matrix. This can be solved as:
\begin{align}
\bm{y}(i) = e^{\bm{\mathcal{K}}i}\bm{y}(0) = \bm{W}e^{\bm{\Omega} i}\bm{W}^{\dagger}\bm{y}(0) \qquad \forall i \in \mathbb{R}\label{eq:koopman-linearize}
\end{align}
The Koopman matrix characterizes the system and contains information about it; in particular, its eigenvectors $\bm{W}$ are referred to as the \ac{DMD} modes, and the associated eigenvalues $\bm{\Omega}$ govern how the system behaves as it is evolved. Linearization is an incredibly powerful technique since any unknown state of a system can be easily computed from any of its known states using well-developed linear techniques, as done above to compute the unknown $\bm{y}(i)$ state from the initial state $\bm{y}(0)$.

In the discrete sampled equivalent of the above, the system is described as $\bm{y}_{i+1} = \bm{K}\bm{y}_i$, where $\bm{K}$ is the Koopman matrix. This can be solved as:
\begin{subequations}
\begin{align}
\bm{y}_i &= \bm{K}^i\bm{y}_0 &\forall i \in \mathbb{Z}\label{eq:koopman-disc-linearize-a} \\
&= \bm{W}\bm{\Lambda}^i\bm{W}^{\dagger}\bm{y}_0 &\forall i \in \mathbb{Z}
\end{align}
\end{subequations}
The eigenvalues $\bm{\Lambda}$ from the discrete case are related to the general eigenvalues $\bm{\Omega}$ as $\bm{\Lambda} = e^{\bm{\Omega}\Delta i}$, where $\Delta i$ is the sampling interval for discretization.

The encoded domain is finally decoded to obtain states in the original domain:
\begin{align}
\bm{x} = g^{-1}\left(\bm{y}\right) \label{eq:koopman-decoder}
\end{align}

\subsection{Highlights of \package{}}\label{sec:intro-highlights}
The core components of our \package{} package are a) an autoencoder architecture that learns an encoder $g(\cdot)$ suited to linearization, and its corresponding decoder $g^{-1}(\cdot)$, and b) a method to learn the Koopman matrix. The latter can be achieved in two different ways, which also correspond to the two primary tasks the package performs -- \emph{state prediction} and \emph{trajectory prediction}.

\subsubsection{State Prediction}\label{sec:intro-highlights-sp}
The goal here is to learn the dynamics of a system from its known states, then predict the values of its unknown states. As an example application, Sec. \ref{sec:dk-examples-sp} describes \package{} performing state prediction of the pressure on the surface of an airfoil at unknown angles of attack.

The input data is $\left\{\bm{x}_i, i\in I\subset\mathbb{R}\right\}$, where $I = \{i_0, i_1, \cdots, i_m\}$ is a set of indexes at which the states of the system have been measured. These input states are encoded to $\{\bm{y}_i\}$, and grouped into matrices $\bm{Y}_{\text{prev}} = \left[\bm{y}_{i_0}\cdots\bm{y}_{i_{m-1}}\right]$ and $\bm{Y}_{\text{next}} = \left[\bm{y}_{i_1}\cdots\bm{y}_{i_m}\right]$. The Koopman matrix can be computed as $\bm{K} = \bm{Y}_{\text{next}}\bm{Y}_{\text{prev}}^{\dagger}$, which uses \ac{SVD}. The subsequent workflow involves computing its eigendecomposition $\bm{W}$ and $\bm{\Lambda}$, using $\bm{\Lambda}$ to compute $\bm{\Omega}$, then using Eq. \eqref{eq:koopman-linearize} to compute $\left\{\bm{y}_{i'}, i'\in I'\subset\mathbb{R}, I'\cap I = \emptyset\right\}$. Finally, these are decoded to get $\{\bm{x}_{i'}\}$, which are the predicted states of the original system at indexes $\{i'\}$ where they were not measured. (Values of $i'$ can be positive, negative, or fractional, which respectively correspond to forward extrapolation, backward extrapolation, or interpolation.) All of these steps are included in the deep learning pipeline.

\subsubsection{Trajectory Prediction}\label{sec:intro-highlights-tp}
Classical applications of Koopman theory provide data in the form of trajectories, i.e. `rollouts' of a system from an initial state for a fixed number of indexes into the future. The goal of trajectory prediction is to learn the dynamics of the system from a given number of known trajectories, then predict unknown trajectories starting from new initial states.

The input data is $\left\{\left[\bm{x}^j_0, \bm{x}^j_1, \cdots, \bm{x}^j_m\right], j\in\{j_1,j_2,\cdots,j_J\}\right\}$ -- the sequence inside square brackets is a trajectory $j$ starting from initial state $\bm{x}^j_0$ and rolled out up till state $\bm{x}^j_m$; there are $J$ such given trajectories. The pipeline begins by encoding all states in each trajectory. While the Koopman matrix can be computed using \ac{SVD} as before, it can be slow for lengthy trajectories. Hence, for trajectory prediction, the \package{} package models the Koopman matrix $\bm{K}$ as the weights of a linear \ac{NN} layer -- a \ac{MLP} with equal number of input and output neurons, no bias, and no activation function or other source of nonlinearity. This can be directly applied to evolve the system instead of performing the eigendecomposition. A linear layer incurs the significant limitation of not being able to predict states for negative or fractional indexes since it cannot be applied backwards or a fractional number of times, however, this limitation is irrelevant to trajectory prediction since trajectory indexes are only integral and always advance forward. Then, given new initial states $\left\{\bm{x}^{j'}_0, j'\notin\{j_1,j_2,\cdots,j_J\}\right\}$, the trained linear layer $\bm{K}$ can be used to generate complete trajectories $\left\{\left[\bm{x}^{j'}_0, \bm{x}^{j'}_1, \cdots, \bm{x}^{j'}_m\right]\right\}$ for each of them using Eq. \eqref{eq:koopman-disc-linearize-a}.

\subsection{Related Work}\label{sec:intro-related}
We first discuss other general-purpose software packages in literature that implement Koopman theory. The Python package \texttt{PyKoopman} (\cite{pykoopman}) -- built using the \texttt{PyDMD} package (\cite{pydmd}) for performing \ac{DMD} only -- can do state and trajectory prediction. However, the user needs to provide a ready-made encoding $\bm{y}$ as input. In the absence of such an input, the encoder used is an identity function, i.e. $\bm{y}=\bm{x}$, wherein Koopman theory reduces to \ac{DMD} only. The Python package \texttt{pykoop} (\cite{pykoop}) allows the user to use particular functions for encoding $\bm{x}$ into $\bm{y}$, such as radial basis functions. A related example is the Matlab toolbox \texttt{koopman} (\cite{mbudisic_koopman}), which constructs $\bm{y}$ as the Fourier Transform of $\bm{x}$. We note that these packages impose a certain amount of restriction regarding how the encoded states are derived from the input states.

The motivation to learn encoded states unrestrictedly with sole focus on achieving good linearization led to the development of deep learning models such as autoencoders to obtain $\bm{y}$ from $\bm{x}$. There have been other efforts along these lines, generally built to train on input trajectories and perform trajectory prediction. While code exists for some of these efforts, such code is usually bespoke and serves to demonstrate the specific results in the respective papers
.

\cite{Lusch2018,Champion2019} focus on finding parsimonious / sparse representations for dynamical systems using autoencoders. In particular, \cite{Lusch2018} learns the Koopman matrix using a linear \ac{NN} layer, and includes an auxiliary network to learn continuous eigenvalues. \cite{Lago2022} created the deep learning \ac{DMD} (DLDMD) effort that has some similarity to our work, but lacks code. \cite{Takeishi2017} uses linear-delay embedding for time-series data. \cite{Azencot2020} uses recurrent \acp{NN} to learn consistent dynamics, similar to \cite{Otto2019}'s \ac{LRAN} to learn low-dimensional encodings, while \cite{Geneva2022} uses transformers. \cite{Otto2019,Williams2015} learn the elements of the Koopman matrix directly, following which they perform its eigendecomposition. \cite{Yeung2017,Li2017} use feed-forward networks to encode the original data, but they do not convert back to the original input domain using a decoder. In particular, \cite{Yeung2017} uses \ac{NN} layers to learn the Koopman matrix, while \cite{Li2017} learns it directly. \cite{Mardt2018,Wehmeyer2018} use deep learning for Koopman theory applied to the specific domain of molecular kinetics. Finally, note that the very recent work of \cite{Lew2023_software} discusses how numerical differentiation may overcome the limitation of linear layers being restricted to predicting positive integral indexes only. However, we reiterate that learning the Koopman matrix directly, as discussed in Sec. \ref{sec:intro-highlights-sp}, completely overcomes the problem by generalizing prediction to any real-valued index.

\section{The \package{} package}\label{sec:dk}
This section describes our core contribution -- \package{}. The \package{} Python package is available on PyPI and can be installed via \code{pip install dlkoopman}. The current version is \code{1.1.2} at the time of final submission, and can be cited using \cite{Dey2023_dlkoopman}. Source code is available at \url{https://github.com/GaloisInc/dlkoopman}. The \code{README} gives a broad overview of the package and links to tutorials. The complete API reference and documentation can be found at \url{https://galoisinc.github.io/dlkoopman/}.

\subsection{Training}\label{sec:dk-train}

\begin{figure}[!ht]
    \centering
    \includegraphics[width=\linewidth]{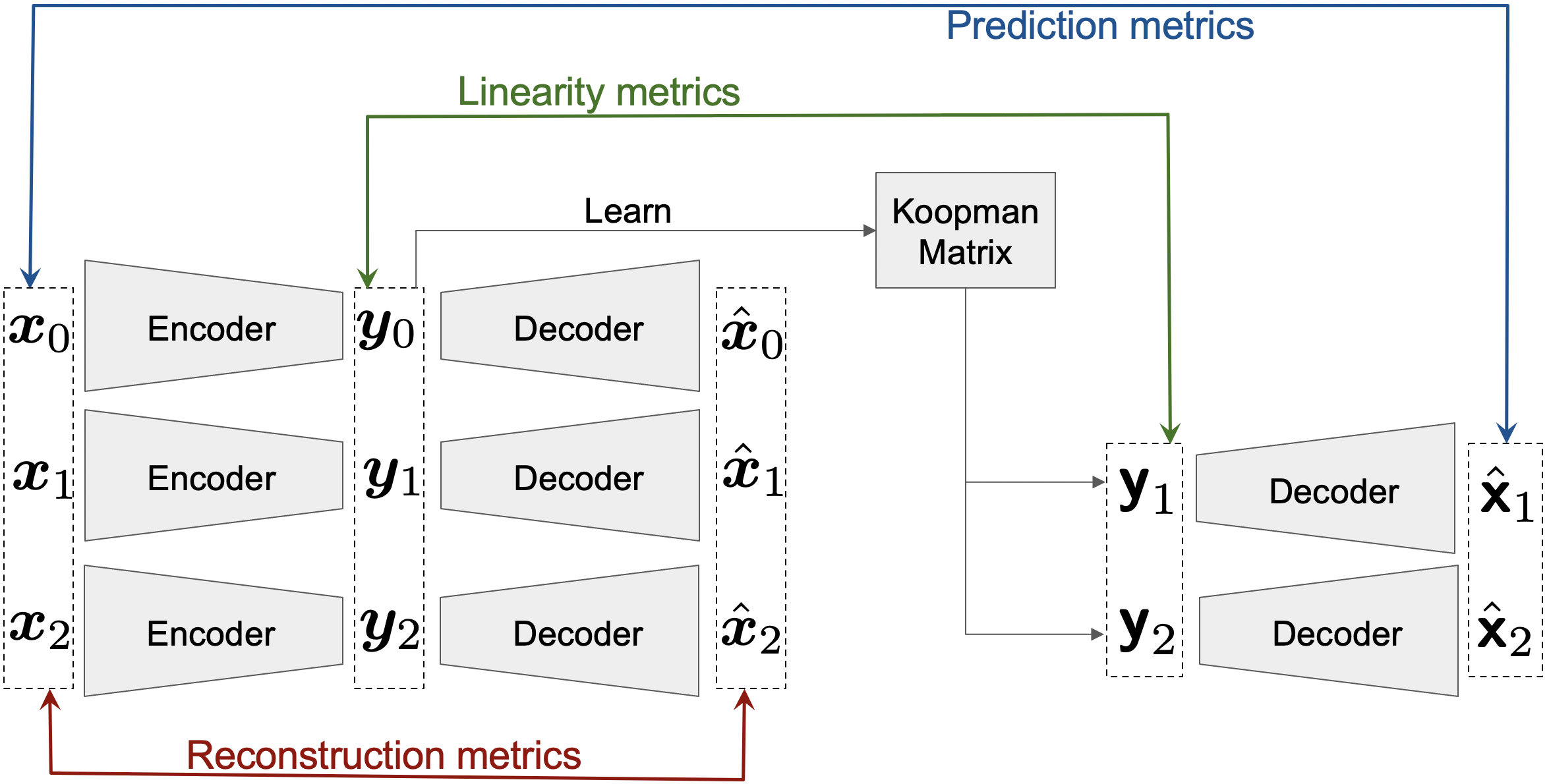}
    \caption{The training algorithm for a small example with three input states $\left[\bm{x}_0, \bm{x}_1, \bm{x}_2\right]$, which can be input to either \SP{} or \TP{} (trajectory superscript omitted in figure). These are passed through an encoder to get encoded states $\left[\bm{y}_0, \bm{y}_1, \bm{y}_2\right]$. These are passed through a decoder to get $\left[\hat{\bm{x}}_0, \hat{\bm{x}}_1, \hat{\bm{x}}_2\right]$, and also used to learn the Koopman matrix. This is used to derive predicted encoded states $\left[\bm{\mathsf{y}}_1, \bm{\mathsf{y}}_2\right]$, which are passed through the same decoder to get predicted approximations $\left[\hat{\bm{\mathsf{x}}}_1, \hat{\bm{\mathsf{x}}}_2\right]$ to the original input states. Training proceeds by minimizing errors -- reconstruction between $\left\{\bm{x}\right\}$ and $\left\{\hat{\bm{x}}\right\}$, linearity between $\left\{\bm{y}\right\}$ and $\left\{\bm{\mathsf{y}}\right\}$, and prediction between $\left\{\bm{x}\right\}$ and $\left\{\hat{\bm{\mathsf{x}}}\right\}$.}
    \label{fig:training_architecture}
\end{figure}

The two core modules of the package are \SP{} for state prediction, and \TP{} for trajectory prediction. They have a similar algorithm during the training phase, the aim of which is to optimize three metrics -- reconstruction, linearity, and prediction. These are standard metrics found in Koopman theory literature (\cite{Lusch2018,Lago2022}). The training algorithm is described in Fig. \ref{fig:training_architecture}.

The input data consists of states $\{\bm{x}\}$ of the system that we wish to model. \SP{} requires the states to be accompanied by indexes, e.g. $\left\{\bm{x}_{10}, \bm{x}_{15}, \bm{x}_{19.5}, \cdots\right\}$ (for the purposes of training, indexes are rounded to have equal spacing, i.e. $\bm{x}_{19.5}$ will be converted to $\bm{x}_{20}$). \TP{} does not require indexes since the states are assumed to form a trajectory $\left[\bm{x}^j_0,\bm{x}^j_1,\bm{x}^j_2,\cdots\right]$. These states are passed through a \ac{MLP} \acl{NN} -- the encoder -- to obtain encoded states $\{\bm{y}\}$. The user needs to provide the dimensionality of the encoded states via the required argument \code{encoded\_size}. The encoded states are passed through a \ac{MLP} -- the decoder -- to obtain reconstructed states $\left\{\hat{\bm{x}}\right\}$. The \emph{reconstruction error} is computed between $\{\bm{x}\}$ and $\left\{\hat{\bm{x}}\right\}$, minimizing this ensures the decoder is learning the inverse function of the encoder.

For \SP{}, the encoded states are used to perform \ac{SVD} to obtain the Koopman matrix. Its eigendecomposition is used to generate predictions $\left\{\bm{\mathsf{y}}\right\}$ (note the different font) for encoded states by using Eq. \eqref{eq:koopman-linearize}. An important consideration here is the \emph{rank} of the \ac{SVD}. Using a lower rank usually results in approximations that generalize better by reducing overfitting, however, a rank too low will incur bias. A lower rank also reduces the probability of numerical issues when performing backpropagation (\cite{pytorch_svd,pytorch_eig}). The user needs to provide \code{rank} since it's a required argument. For \TP{}, the initial encoded state $\bm{y}_0$ is evolved via a linear layer, which serves as the Koopman matrix. This can be used to generate the entire predicted trajectory $\left[\bm{\mathsf{y}}_1,\bm{\mathsf{y}}_2,\cdots\right]$ by using Eq. \eqref{eq:koopman-disc-linearize-a}. The \emph{linearity error} is computed between $\{\bm{y}\}$ and $\left\{\bm{\mathsf{y}}\right\}$, minimizing this ensures a good Koopman matrix is being learnt that can achieve linearization.

The predictions for encoded states / trajectories are decoded to obtain predictions $\left\{\hat{\bm{\mathsf{x}}}\right\}$ (note the different font) for input states / trajectories. The \emph{prediction error} is computed between $\{\bm{x}\}$ and $\left\{\hat{\bm{\mathsf{x}}}\right\}$, minimizing this ensures a good overall pipeline that learns the autoencoder and Koopman matrix.

\subsection{New predictions}\label{sec:dk-pred}

\begin{figure}[!ht]
    \centering
    \includegraphics[width=0.8\linewidth]{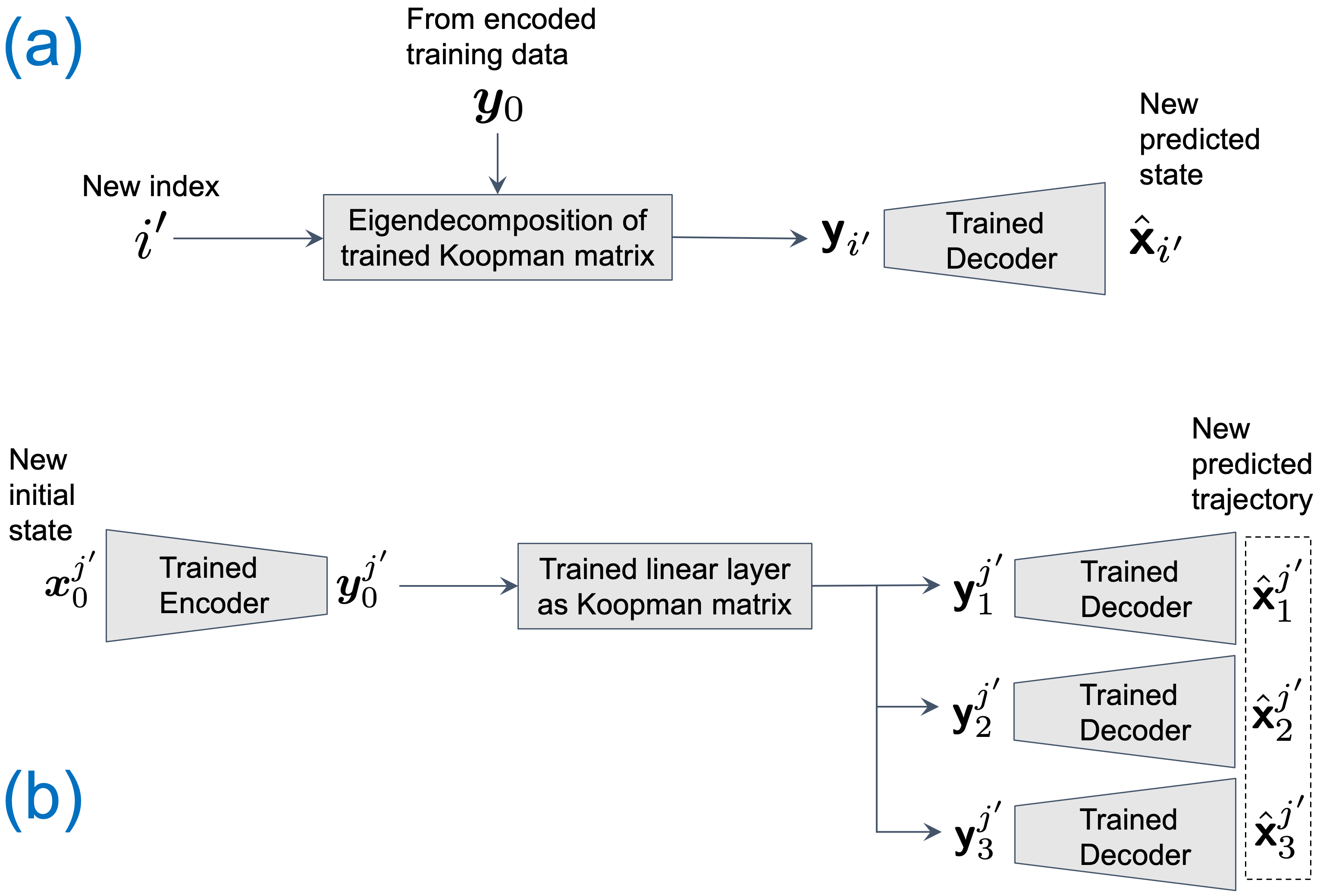}
    \caption{After training, (a) the \SP{} can be used to compute predicted states for new indexes such as $i'$, (b) the \TP{} can be used to generate predicted trajectories for new starting states such as $\bm{x}^{j'}_0$.}
    \label{fig:prediction_architecture}
\end{figure}

New predictions can be computed after training finishes. For \SP{}, given any new index $i'$ not present in the input data, the architecture can predict $\hat{\bm{\mathsf{x}}}_{i'}$, as shown in Fig. \ref{fig:prediction_architecture}(a). For \TP{}, given any new initial state $\bm{x}^{j'}_0$ not present in the input data, the architecture can predict the trajectory $j'$ as $\left[\hat{\bm{\mathsf{x}}}^{j'}_1,\hat{\bm{\mathsf{x}}}^{j'}_2,\cdots\right]$, as shown in Fig. \ref{fig:prediction_architecture}(b).

\subsection{Metrics}\label{sec:dk-metrics}
As described in Sec. \ref{sec:dk-train}, three metrics are computed to judge the goodness of a training run -- reconstruction, linearity, and prediction. The overall loss function is:
\begin{align}
L = L_{\text{lin}} + \alpha\left(L_{\text{recon}}+L_{\text{pred}}\right) + \beta L_{\text{Autoencoder}} + \gamma L_{\bm{K}} \label{eq:loss}
\end{align}
This overall loss $L$ is optimized when performing backpropagation to train the deep learning architecture. $L_{\text{lin}}$, $L_{\text{recon}}$ and $L_{\text{pred}}$ are computed using \ac{MSE}. $\alpha$ is a term that is used to weight the losses for quantities that are outputs from the decoder, i.e. reconstruction and prediction loss. $\beta$ is the coefficient of weight decay $L_{\text{Autoencoder}}$ for the autoencoder parameters. $\gamma$ is the coefficient of a penalty $L_{\bm{K}}$ for the elements of the Koopman matrix. For \SP{}, $L_{\bm{K}}$ is the average absolute value of the elements of $\bm{K}$, while for \TP{} -- where $\bm{K}$ is a linear layer -- $L_{\bm{K}}$ is weight decay for the parameters of the linear layer, and $\beta = \gamma$.


While losses are suitable for optimizing during gradient descent, they are dependent on the scale of the data. This makes them unsuitable for human-readability in terms of how well a \SP{} run is performing. This motivates us to introduce a novel metric \acf{ANAE}, which can be defined between two tensors of same size as:
\begin{align}
\text{ANAE}(\bm{p},\bm{q}) = \text{Avg}_{p_i\ne0}{\left(\frac{\left|p_i-q_i\right|}{\left|p_i\right|}\right)}\label{eq:anae}
\end{align}
where $\bm{p}$ is the reference, and $\bm{q}$ is the prediction. Note that $\bm{p}$ and $\bm{q}$ can have any number of dimensions, they will be flattened before computing \ac{ANAE}.

Thus, \ac{ANAE} tells us how far off we can expect each element of the prediction to be from the corresponding element in the reference. As an example, consider the reference data to have two 3-dimensional states: $[-0.1,0.2,0]$ and $[100,200,300]$. Suppose a trained \SP{} model produces predictions $[-0.11,0.15,0.01]$ and $[105,210,285]$ for this reference data. Normalizing the absolute deviations by the absolute reference values yields $10\%$ and $25\%$ for the first two values, the third value is ignored since its reference value is $0$, and the last three values are each $5\%$. Taking the average of these yields $10\%$. This is a useful figure. It tells us that when predicting unknown states for which reference values are not available, the user can expect the current model to predict values that are $\sim10\%$ off. Note that just like loss, \ac{ANAE} can also be computed for reconstruction, linearity, and prediction. Possibly the most important out of these is the prediction \ac{ANAE}, since it tells us the error we can expect from unknown state prediction.

\subsection{Example Applications}\label{sec:dk-examples}
The source code for \package{} includes complete tutorials in the \code{examples/} folder.

\subsubsection{State Prediction}\label{sec:dk-examples-sp}
Here, we briefly describe an application of the \SP{}. This example attempts to predict the pressure vector across the surface of a NACA0012 airfoil at varying angles of attack -- a commonly studied \ac{CFD} problem (\cite{Critzos1955,Voodarla2021}). A particular state $\bm{x}_i$ for this case would be the $200$-dimensional pressure vector $\bm{x}$ at angle of attack $i$.

The data consists of pairs of variables prefixed by \code{X} for the states, and \code{t} for the indexes. (Note that the code uses prefix \code{t} for indexes since \code{i} is a variable commonly reserved for iterators.) Three such pairs may be provided -- \code{Xtr} and \code{ttr} form the training data that must be provided, while \code{Xva} and \code{tva}, and \code{Xte} and \code{tte}, respectively, are validation and test data that may be optionally provided to check the performance of training. For this example, each of the \code{t} variables is a list containing values of angles of attack, while each corresponding \code{X} variable is a matrix of dimensions (\code{length(t)}$\times\,200$) where each row contains the pressure vector for a particular angle of attack.

\begin{figure}[!ht]
    \centering
    \includegraphics[width=\linewidth]{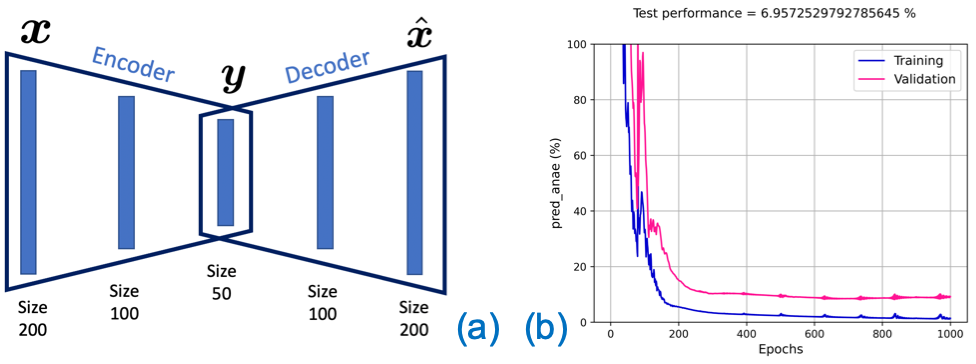}
    \caption{(a) Example autoencoder architecture. Dimensionality of the input states $\bm{x}$ and reconstructed states $\hat{\bm{x}}$ is $200$, and that of the encoded states $\bm{y}$ is \code{encoded\_size=50}. The encoder is set to have one hidden layer of dimensionality \code{encoder\_hidden\_layers=[100]}. The decoder here is the mirror image of the encoder, but it can be configured to be otherwise. (b) Example results for prediction \ac{ANAE} after training a \SP{} model on NACA0012 airfoil pressure data at varying angles of attack. The plots show prediction \ac{ANAE} on training and validation data from epochs 100 to 1000, and final prediction \ac{ANAE} on test data is noted at the top.}
    \label{fig:naca0012_combined}
\end{figure}

A \SP{} instance can be created as:
\begin{minted}{python}
sp = StatePred(
    dh = dh, #StatePredDataHandler instance used to pass data
    rank = 6,
    encoded_size = 50,
    encoder_hidden_layers = [100]
)
\end{minted}
which results in an autoencoder that looks like Fig. \ref{fig:naca0012_combined}(a). This can be trained (and validated, if \code{Xva} and \code{tva} are provided) as:
\begin{minted}{python}
sp.train_net(
    numepochs = 1000,
    decoder_loss_weight = 0.1, #alpha in the loss equation
    weight_decay = 1e-5, #beta in the loss equation
    Kreg = 0 #gamma in the loss equation
)
\end{minted}

An epoch of training comprises computing the Koopman matrix using the entire training data, using it to get predictions, then computing metrics on training data, and validation data if given. If \code{Xte} and \code{tte} are provided, test statistics may be obtained after training via \code{sp.test\_net()}.

The \code{utils} module of the package contains a utility to plot loss and \ac{ANAE} statistics. An example plot is shown in Fig. \ref{fig:naca0012_combined}(b), which achieves $6.95\%$ prediction \ac{ANAE} on test data.

Now the pressure vector can be predicted at unknown angles of attack that were not present in either of \code{ttr}, \code{tva} and \code{tte}. As examples, we pick an \emph{interpolated} index $3.75^{\circ}$, and an \emph{extrapolated} index $21^{\circ}$. The predictions are yielded by running:
\begin{minted}{python}
sp.predict_new([3.75,21])
\end{minted}

\subsubsection{Trajectory Prediction}\label{sec:dk-examples-tp}
The primary difference in a \TP{} run compared to a \SP{} run is that each epoch is sub-divided into batches, with each batch containing a subset of training data trajectories. The starting index of each of these trajectories is evolved via the linear layer to get predictions for the rest of each trajectory, and compute training metrics. Validation metrics are computed at the end of all batches in an epoch, while test metrics are computed at the end of all epochs, as before.

The versatility of the \package{} package allows it to run on different kinds of data, which naturally also includes data sets that have been used in prior literature. The source code contains an example of running \TP{} on data from a system exhibiting a polynomial manifold, described by the equations $\dot{x}_1 = \mu x_1$ and $\dot{x}_2 = \lambda\left(x_2-x_1^2\right)$, with $\lambda<\mu<0$. This has been studied previously in \cite{Brunton2016} and \cite{Lusch2018}, and we use data from the latter. We trained a \TP{} model on $\sim10000$ trajectories; the results are in the \code{examples/} folder of the source code and have been omitted here for brevity considerations.

\subsection{Hyperparameter search}\label{sec:dk-hypsearch}
Both the \SP{} and \TP{} contain several arguments / hyperparameters that affect the architecture and training pipeline. A full list of these can be obtained from the API reference. Setting suitable values for these hyperparameters is important to achieving good performance. While there are techniques in literature such as \cite{AutoPytorch,DeepnCheap} that automatically search for the best model to use on given data, they may be challenging to adopt to \package{}. To ease this burden, we provide a hyperparameter search module in the \package{} package that is ready to be used for \SP{} and \TP{}. The key idea is that the user can provide ranges for each hyperparameter to be swept over. Each set of hyperparameter values leads to a particular model; many such models are then run on the given data and the results compiled and ranked, from which the user can choose the best model(s).

The \code{examples/} folder in the source code contains tutorials for hyperparameter search. Here, we briefly run through an example. Suppose the user provides:
\begin{minted}[breaklines]{python}
hyp_options = {
    'rank': [3,4,5,6], #4 options
    'encoded_size': 50, #1 option
    'encoder_hidden_layers': [[200,200],[300,200,100]], #2 options
    'numepochs': [200, 300, 400] # 3 options
}
\end{minted}

This will result in a total of $4\times1\times2\times3=24$ sets of hyperparameter values, which will use different values of \code{rank}, \code{encoder\_hidden\_layers}, and \code{numepochs}. All other arguments will be set to the provided values (i.e. \code{encoded\_size = 50}), or defaults if not provided. Hyperparameter search can then be run as:
\begin{minted}[breaklines]{python}
run_hyp_search(
    dh = dh, # instance of either StatePredDataHandler or TrajPredDataHandler, depending on the problem
    hyp_options = hyp_options,
    numruns = 15, # optional
    sort_key = 'avg_pred_anae_va' # optional
)
\end{minted}

Since a model can take time to run for large data sets, the optional argument \code{numruns} allows the user to specify how many runs they want. In this case, 15 models will be sampled from the total of 24 possibilities. The optional argument \code{sort\_key} defines the metric to sort the final results by. In this case, the average prediction \ac{ANAE} on validation data will be used to rank the models. We strongly recommend providing validation data via the data handler, as performance on validation data is a good indicator of the goodness of any model.

Finally, note that if the script is halted (such as the user stopping it forcibly, or other unforeseen interruptions), intermediate results will be available. This is particularly useful for long runs.

\subsection{Miscellaneous notes}\label{sec:dk-misc}
The speed of executing individual runs depends on a variety of factors such as the amount of training data, complexity of the autoencoder network, and the computing platform. \package{} uses Pytorch, and will run on GPUs by default if available. The user can change this default, as well as several other configuration options such as the choice to use exact eigenvectors vs projected eigenvectors for state prediction (\cite{DMDbook}). These configuration options have been included to give freedom to the user -- they are described under \code{dlkoopman.config} in the documentation.

\section{Conclusion}\label{sec:conc}
We have presented \package{} -- a software package for Koopman theory that uses a deep learning autoencoder to learn linear encodings of a system, along with learning the corresponding linear dynamics. It trains on data from snapshots of any system, and can perform both state prediction and trajectory prediction. The package is open-source and can be installed as a Python tool. It contains extensive documentation, examples, and a hyperparameter search module, and introduces a novel performance metric \ac{ANAE}. We hope that \package{} benefits many and becomes widely used as a tool for data-driven analysis and behavioral predictions of dynamical systems. Future work will include adding features such as control inputs and additional losses.

\acks{%
This material is based upon work supported by the United States Air Force and DARPA under Contract No. FA8750-20-C-0534. Any opinions, findings and conclusions or recommendations expressed in this material are those of the author(s) and do not necessarily reflect the views of the United States Air Force and DARPA. Distribution Statement A, ``Approved for Public Release, Distribution Unlimited.''

The authors would like to thank Ethan Lew and Pachapakesan Shyamshankar for helpful technical discussions and insights, and Matt Le Beau for administrative help.
}

\bibliography{aaa_main}

\end{document}